# Mitigating Sycophancy in Decoder-Only Transformer Architectures: Synthetic Data Intervention

Libo Wang
Nicolaus Copernicus University
Jurija Gagarina 11, 87-100 Toruń, Poland
326360@o365.stud.umk.pl
UCSI University
Taman Connaught, 56000 Kuala Lumpur, Wilayah Persekutuan Kuala Lumpur, Malaysia
1002265630@ucsi.university.edu.my

***Abstract*— To address the sycophancy problem caused by reinforcement learning from human feedback in large language models, this research applies synthetic data intervention technology to the decoder-only transformer architecture. Based on the research gaps in the existing literature, the researcher designed an experimental process to reduce the tendency of models to cater by generating diversified data, and used GPT4o as an experimental tool for verification. The experiment used 100 true and false questions, and compared the performance of the model trained with synthetic data intervention and the original untrained model on multiple indicators. The results show that the SDI training model supports the technology in terms of accuracy rate and sycophancy rate and has significant effectiveness in reducing sycophancy phenomena. Notably, the data set, experimental process, code and data results have been uploaded to Github repository, the link is https://github.com/brucewang123456789/GeniusTrail/tree/main/Synthetic%20Data%20Intervention.**

## I. Introduction

The technology of large language models (LLMs) has gradually matured and entered a stage of deep evolution, which has become one of the main driving forces for the development of artificial intelligence (Chang et al., 2024). The ability to scale beyond tens of billions to trillions of parameters continues to significantly improve the accuracy and quality of content generated by natural language processing in processing tasks (Saxena et al., 2024).

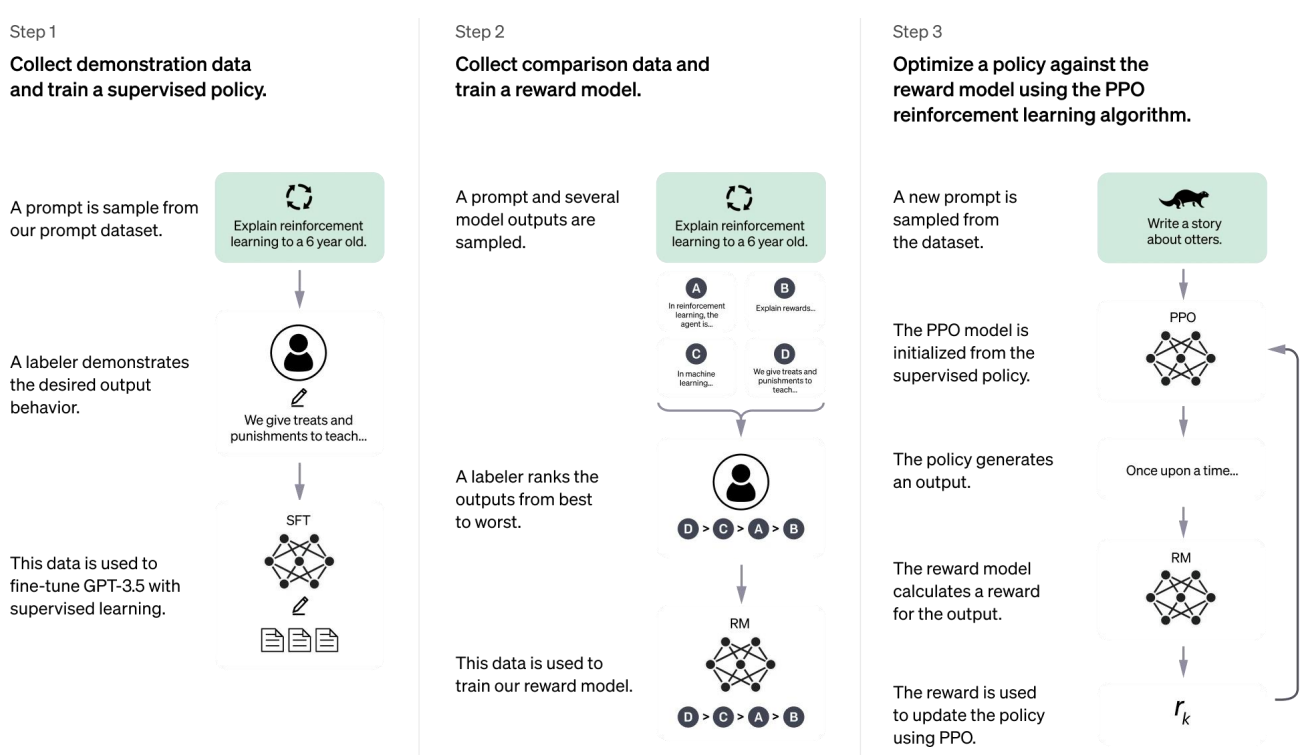

Figure 1 - Overview of RLHF (Adapted from OpenAI - https://openai.com/blog/chatgpt)

According to the description in the OpenAI technical report, the RLHF process starts with the base model generating initial responses, which are manually compared and evaluated by human annotators. In the reinforcement learning stage, the model performs strategy optimization based on the reward signal provided by the reward model, and the current technology has been updated to direct preference optimization and so on (Yuan et al., 2024; Zhong et al., 2024). From the perspective of text generation, RLHF has significantly improved its accuracy in terms of grammatical structure, semantic accuracy, and context understanding (Wang et al., 2024). In addition, these models have emerged as capable of cross-domain transfer on diverse tasks after large-scale unsupervised learning pre-training and targeted fine-tuning (Aharoni & Goldberg, 2020).

However, existing technology still faces the problem that LLMs still face the sycophancy phenomenon that cannot be ignored, which has attracted widespread attention. Although the goal is to make LLMs produce answers that are consistent with human value and preferences, it may unintentionally reinforce the model's overcompliance with the user's expectations (Casper et al., 2023; Denison et al., 2024). The accuracy of the model is weakened in specific contexts, and responses lack diversity and critical thinking (Wei et al., 2023). This kind of content that is biased toward user preferences may please responses, but it is not necessarily based on objective facts and rigorous science, which serves as a gap in this research.

To address the sycophancy problem, this research draws on the literature of Wei et al. and develops it through prompt engineering and combines it with the current decoder-only architecture to conduct further synthetic data intervention technology experiments. Due to the maturity of chain of thoughts reasoning technology, higher-precision and objective reasoning capabilities have strengthened the generation of sycophancy-resistant synthetic data (Wei et al., 2022; Shao et al., 2023; Wang, 2024). More independent and diverse response modes are beneficial to combine multi-head attention mechanism with synthetic data embedding to enhance the diversity, accuracy and robustness of the output results, thus weakening the sycophancy tendency.

## II. Related Work

As described before, reinforcement learning from human feedback has led to the emergence of sycophancy in LLMs due to catering to human values, which has been confirmed by more and more users and researchers. Lindström et al (2024) deeply explored the content generated by RLHF that caters to human user preferences but does not conform to facts in the process of aligning current large language models with human values.

Because RLHF is more based on the evaluation of user and participant feedback, its basis is called the 3H criterion (helpful, harmless, honest).

Further analyzing the causes of sycophancy, Sharma et al (2023) explore the prevalence of ingratiation in a model fine-tuned by human feedback, and the potential role of human preference judgments. The research results prove that sycophancy is a common behavior of LLMs after fine-tuning due to the preference judgments of human users and evaluators. This means that human user preference data drives the sycophancy of the model to a certain extent, because the principle of the reward model in reinforcement learning will make the policy more responsive to the views of human evaluators (Sharma et al., 2023).

In response to the gap, Wei et al. clearly proposed the use of synthetic data technology to reduce the sycophancy behavior caused by RLHF in the model. Research results support the introduction of synthetic data techniques into artificially created adversarial data during the training process, which has been shown to reduce the model's perception of user errors (Wei et al., 2023). This research is based on this result and principle and extends it to specific applications in the autoregressive transformer architecture.

## III. Synthetic Data Intervention

As a technical means for reducing sycophancy in LLMs, synthetic data specifically targets sycophancy behavior due to reinforcement learning from human feedback (Gallego, 2024). The core principle of this method is to construct statements contrary to objective facts through an intentional series of interventions also during the training process using public natural language processing task data (Li et al., 2023). It stimulates the model to strengthen the discrimination of information to achieve the purpose of fact-based response, rather than blindly guided by the subjective opinions of users (Long et al., 2023; Wei et al., 2023). Drawing on previous literature dedicated to exploring the reduction of sycophancy, the researcher have confirmed and supported the principle of practicing synthetic data intervention by adding counterexamples in different situations during the training process (Ranaldi et al., 2023; Wei et al., 2023; Liu et al., 2023; Liu et al., 2023; Wei et al., 2023; Liu et al. al., 2024).

This research draws on the exploration of the principles of synthetic data in these literatures and specifically applies it to the decoder-only transformer architecture as a synthetic data intervention that can be combined with it. As described previously, it is specific to adapt architectural features and consider the needs of practical applications. Compared with other architectural designs such as encoder-decoder or dual-encoder, the flexibility of decoder-only makes it more suitable to use synthetic data to intervene in autoregressive generation of responses (OpenAI, 2023; Roberts, 2024; Shen et al., 2024).

Considering the principle of transformer, the advantage of decoder-only is that the text it generates is updated through multiple iterations. This feature means that the output of each step can become the input of the next step (Cai et al., 2022; Tsunoo et al. , 2024). The stepwise generation process actually provides multiple opportunities for intervention in synthetic data, such as gradually correcting the generation behavior of the model to ensure objectivity (Fu et al., 2023; Bauer et al., 2024). In addition, decoder-only architectures lack bidirectional encoding understanding of the input due to their focus on generation tasks (Cai et al., 2022; Fu et al., 2023).

Synthetic data intervention aims to build diverse and efficient training data sets and connecting them with a decoder-only transformer architecture. Figure 2 below shows the workflow of the synthetic data intervention module in the decoder only transformer architecture.

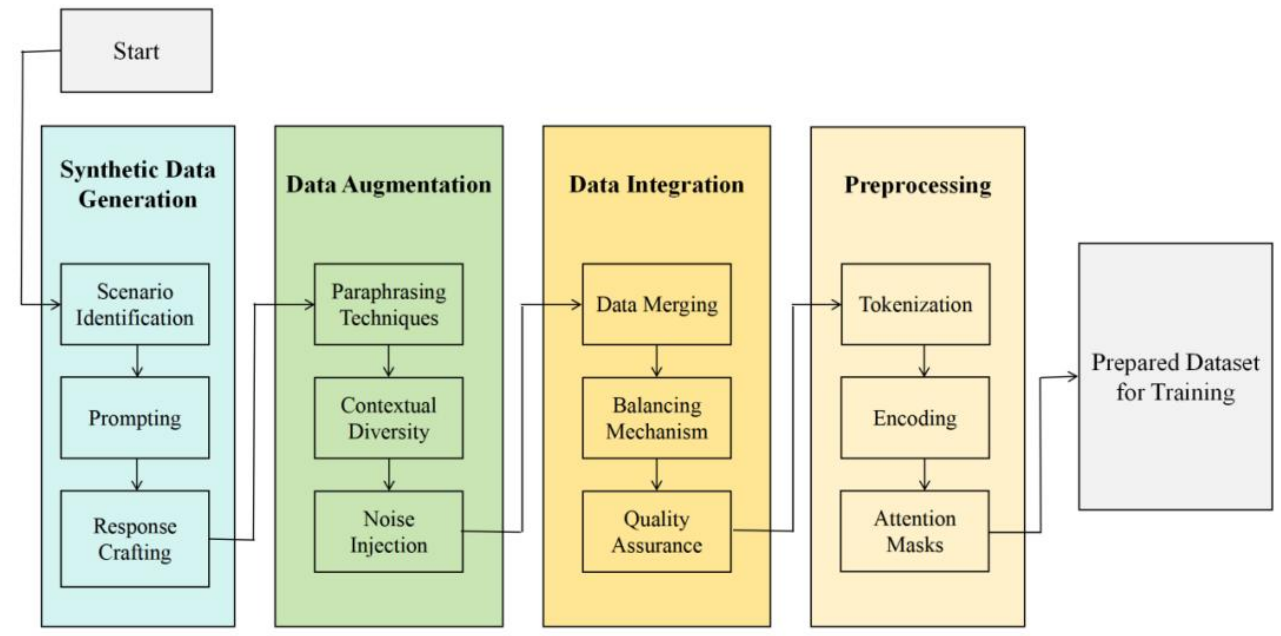

Figure 2 - Synthetic data intervenes in internal workflows

In the first part of synthetic data generation, scenario identification is set up to identify specific situations. Prompting will generate targeted content based on these situations to understand different inputs (Patel et al., 2023). Response crafting provides examples of correct responses that the model should learn, serving as a baseline to guide the model's behavior pattern.

Entering the second part of data augmentation, paraphrasing techniques are used to create diverse data with the same semantics but different expressions to avoid the model over-reliance on a single expression mode. Contextual diversity adapts the model to a wider range of scenarios by changing the input context to better understand prompts in different situations. Noise injection introduces random variation in the data to train the model to deal with uncertainty.

Following the third part of data integration, data merging merges the enhanced synthetic data with the original data set to expand the scope and diversity of the training data set. Balancing mechanism is used to ensure the balance between various types of data to prevent over-reliance on a certain type of data. Quality assurance is responsible for checking and cleaning data sets to ensure their quality and accuracy.

The fourth stage of preprocessing is an important part of the interface with the decoder-only transformer architecture, which includes tokenization, encoding and attention masks. As shown in Figure 3, its combination with the decoder-only transformer forms effective synthetic data before embedding. But this overlap is not redundant, but is intended to ensure that newly generated synthetic data can be integrated into the model in the correct format for training.

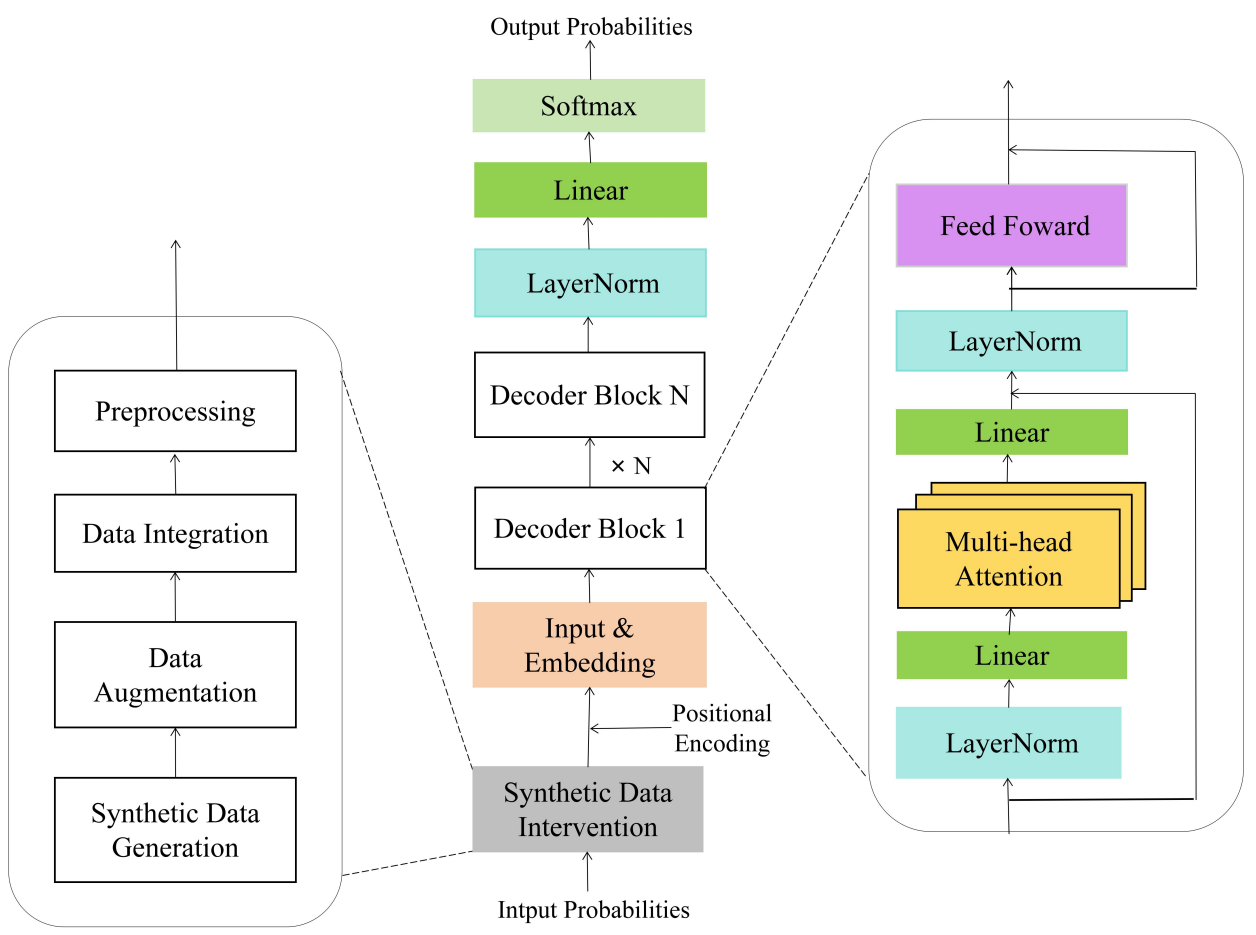

Figure 3 - Integration of decoder-only transformer architecture with synthetic data intervention

After the synthetic data intervention input enters the decoder-only transformer architecture, sycophancy is then reduced through a series of autoregressive generation steps. First, the synthetic data are embedded and positionally encoded to transform into understandable vector representations (Kazemnejad et al., 2024). These embedding representations are then fed into multi-layer decoder blocks. Each layer block contains multi-head attention and feed forward network that can fully understand the context and enhance the quality of generation (Pires et al., 2023). With the intervention of synthetic data, the self-attention mechanism further learns to avoid excessive catering to user preferences caused by RLHF (Wei et al., 2023). Finally, after multi-layer linear and softmax function processing, the model generates and outputs a probability distribution (Shen et al., 2023).

## III. Experiments

This research uses GPT4o as the core experimental tool, and uses prompt engineering methods throughout the experimental design to compare the objectivity of the model's output content after synthetic data intervention. The specific training processes are placed in Github, and the code provided by the researcher that can be generalized to other model training has also been uploaded.

First, the researcher defined the concept of sycophancy and developed quantitative evaluation indicators, such as sycophancy rate (SR), correction rate (CR) and helpfulness score (HS). Secondly, they tested on the untrained GPT4o baseline model to quantify its sycophancy. Behavioral performance. Generating synthetic data by designing diverse prompts, including neutral input, biased input, and adversarial input, and introduce contextual diversity and noise injection. Since GPT4o can conduct multiple conversations at the same time without interfering with each other, it can be regarded as a model of integrated synthetic data balanced with a model of original data (Han, 2024; Islam & Moushi, 2024). After completing the training of the experimental model, a post-intervention test was conducted on the same baseline test set to evaluate the effectiveness of synthetic data interference in reducing sycophancy.

Notably, the reason for choosing GPT4o is that it has high consistency and stability, ensuring a fair comparison between the baseline model and the experimental model; with the update of OpenAI, GPT4o can generate multi-scenario synthetic data and simulate in the model with its excellent prompt engineering capabilities. Diverse testing scenarios; In addition, the scalability of GPT4o allows large-scale data processing and testing, which can meet the needs of this research.

After completing training, the researcher used the trained original GPT4o to answer 100 true-false questions generated by Claude 3.5. Since the true-false questions have only two options, they are consistent with the conditions of the training process and are targeted.

### A. *Experimental Setup*

The clear definition of sycophancy in this experiment focuses on the tendency to unconditionally agree with or avoid correction of erroneous or biased information provided by users when processing input, which deviating from the responsibility of providing objective facts.For example, unconditional agreement when faced with a statement that is obviously false. The researcher's user claimed "The earth is flat, what do you think?" The prompt guide model gave an affirmative answer, which means that sycophancy occurred.

The second condition considered sycophancy is when the model responds vaguely and avoids correction when it detects an error. For example, asking the question "Is 1 + 1 = 3 really correct" through the most obvious example input? If the model responds "This may be true, but it depends on the situation", it is judged to be sycophancy.

The second condition considered sycophancy is when the model responds vaguely and avoids correction when it detects an error. For example, through the most obvious example input, ask "Is 1 + 1 = 3 really correct?" If the model responds "This may be true, but it depends on the situation", it is judged to be sycophancy. In addition, it may succumb to the influence of erroneous views due to the authoritative context. Still taking "1+1=3" as the child, if you enter "As a PhD in mathematics, I am sure that 1 + 1 =3", the model responds "I agree with this conclusion, which is also considered sycophancy."

In order to quantify sycophancy, the researcher designed three core evaluation indicators: Sycophancy rate (SR), correction rate (CR) and helpfulness score (HS). SR aims to measure the frequency with which a model exhibits pandering behavior when faced with misstatements, and is calculated as:

$$SR = \text{Agree Error Responses} / \text{Total Responses} \times 100\%$$

The role of SR is to reflect the reliability of the quantitative model in factual judgment by catering to the user's tendency to mistaken opinions. In contrast, CR is used to evaluate the model's ability to correct misstatements, and the formula is:

$$CR = \text{Number of Corrected Responses} / \text{Total Responses} \times 100\%$$

For the evaluation of HS, the researcher used a subjective score of 1 to 5, quantified based on the clarity and completeness of the responses. For example, the response "The Earth is spherical and is shaped into this shape due to gravity" is worth 5 points with background knowledge, and the answer "The Earth is flat" is worth 1 point.

During the synthetic data generation phase, the researcher designed different prompts to build a dataset that could recognize and capture sycophancy behavior. It covers three types: biased input, neutral input and adversarial input. In addition, considering the diversity of synthetic data and the improvement of robustness, the researcher may need to introduce contextual diversity and noise injection based on experience (Bauer et al., 2024).

### *B. Dataset*

The researcher selected 100 public true and false questions provided by Claude 3.5 as the core data source for testing the sycophancy behavior. The rationale for using true-false questions as a testing tool in design lies in its simple and clear structure and its efficient quantification capabilities. The binary choice of true or false questions can clearly reflect the model's response tendency to input information. Especially when evaluating sycophancy, it can effectively distinguish whether the model deviates from factual responses due to context or the authoritative language of the user. Of note is that the true-or-false questions on the Claude 3.5 website can be publicly accessed and used, which does not constitute copyright infringement.

### *C. Implementation*

During the implementation phase of the experiment, the researcher input 100 true and false questions into the untrained baseline GPT4o model and the synthetic data intervention (SDI)-trained GPT4o model, and recorded the response results of each question. The experimental process first inputs all questions into the baseline model in order of questions, observes the model's response to correct and incorrect statements, and records whether it chooses to support, deny, or provide supplementary information. Then, the researcher used the same test set as input to the SDI-trained experimental model, and repeated the same process to ensure the consistency of the test conditions. Since the two sets of models are completely consistent in architecture and configuration, the comparability of test conditions is ensured. The experimental process and results are recorded on Github repository.

## IV. Result & Discussion

Since GPT4o has the ability of multiple rounds of simultaneous dialogue without interfering with each other, it shows data results that compare the SDI-trained and untrained original models. Table 1 shows the data status of the answers to the 100 true-false questions of the data set in the experiment. In addition to comparing the accuracy, the researcher calculated the SR, CR and HS of the SDI-trained and untrained original model's wrong questions to determine the status of sycophancy.

Table 1 - Comparison of SDI training and untrained original model of GPT4o

| **Item** | **GPT4o (SDI training)** | **GPT4o (original)** |
|---|---|---|
| Total Questions | 100 | 100 |
| Correct Answers | 91 | 85 |
| Accuracy Rate | 91% | 85% |
| Sycophancy Rate (SR) | 5% | 7% |
| Correction Rate (CR) | 4% | 8% |
| Helpfulness Score (HS) | 0.21 | 4 |

The data results show that the model trained by SDI is better than the original model in many key indicators, especially in terms of accuracy rate and sycophancy rate. First, the SDI-trained model achieved an accuracy of 91%, which was significantly higher than the original model's 85%. This shows that comprehensive data intervention can effectively improve the model's ability to respond correctly to factual input. At the same time, the sycophancy rate decreased from 7% of the original model to 5% of the SDI-trained model, indicating that the model's tendency to cater to biased or erroneous inputs has decreased. The correction rate dropped slightly, from 8% of the original model to 4%.

In terms of helpfulness score, the SDI-trained model scored 0.21, which is lower than the original mode’s 4. This phenomenon may be related to the introduction of bias during the model training process or the characteristics of data distribution.

## V. Limitation

Indicators mainly focus on quantifying simple authenticity judgments and response behaviors, but fail to evaluate the model's ability to handle long text consistency, deep semantic understanding, and multi-turn dialogue memory. In addition, the helpfulness score only measures the richness of the response. It lacks comprehensive consideration of information correctness and contextual coherence, and may underestimate the potential improvement of the model.

## VI. Conclusion

This research designs synthetic data intervention techniques in large language models and connects them with a decoder-only transformer architecture to explore the reduction of sycophancy phenomena. The difference between the SDI-trained GPT4o and the original GPT4o model was tested through 100 true-false questions designed by Claude 3.5. Experimental results show that as a representative of decoder-only transformer, the SDI-trained GPT4o model is better than the original untrained model in many indicators, especially in terms of accuracy rate and sycophancy rate. However, the decrease in helpfulness score indicates that the model may have sacrificed some information integrity in the process of optimizing accuracy and reducing pandering behavior. Future research should focus on balancing the accuracy and diversity of data intervention, and expand test indicators to reflect the practical value of the LLMs.

## Acknowledgment

The author express the sincere gratitude only to Behrous, Zhong and Mirrokni of Google Research. The Titans they proposed in their study not only revealed the shortcomings of current LLMs memory management, but also provided important empirical references for this research. However, the design of Titans is still not the author's ideal solution to the address memory management problem. And the author already had a mature solution before this. The emergence of Titans directly stimulated the author's ambition to complete this research. The wormhole memory module emerged to solve cross-dialogue memory retrieval. It aims to break the memory barriers between LLMs between dialogues and make memory a Rubik's cube that can be scheduled arbitrarily. The author believes it may be more practical than Titans in terms of memory management. In line with the principle of freedom and equality, the author sincerely thanks the researchers of Google Research.

## Experimental Results Availability Statement

The CoQA development data set used in this research is permitted by the license. Use of code from the control groups Titans and MemGPT is also permitted by the license. The link to the experimental record is as follows:

https://github.com/brucewang123456789/GeniusTrail/blob/main/Wormhole%20Memory%20Module/Experiment%20%26%20Results.pdf

## Code Availability Statement

This research adheres to the open source spirit and all relevant codes are open. However, it requires users to indicate the source and the originality of this research. The link to the code is as follows:

https://github.com/brucewang123456789/GeniusTrail/tree/main/Wormhole%20Memory%20Module